# FORCE: Feature-Oriented Representation with Clustering and Explanation

Rishav Mukherjee[1+]
Jeffrey Ahearn Thompson[2*+]

Contact information :

Rishav Mukherjee ,MSc., Graduate Research Assistant, Department of Biostatistics & Data Science, University of Kansas Medical Center, Kansas City, Kansas, ZIP 66103, USA

Jeffrey Ahearn Thompson, PhD., Associate Professor, Department of Biostatistics & Data Department of Biostatistics & Data Science, University of Kansas Medical Center, Kansas City, Kansas, ZIP 66103, USA

*jthompson21@kumc.edu

[+] these authors contributed equally to this work

## Abstract

Learning about underlying patterns in data using latent unobserved structures to improve the accuracy of predictive models has become an active avenue of deep learning research. Most approaches cluster the original features to capture certain latent structures. However, the information gained in the process can often be implicitly derived by sufficiently complex models. Thus, such approaches often provide minimal benefits. We propose a SHAP (Shapley Additive exPlanations) based supervised deep learning framework FORCE which relies on two-stage usage of SHAP values in the neural network architecture, (i) an additional latent feature to guide model training, based on clustering SHAP values, and (ii) initiating an attention mechanism within the architecture using latent information. This approach gives a neural network an indication about the effect of unobserved values that modify feature importance for an observation. The proposed framework is evaluated on three real life datasets. Our results demonstrate that FORCE led to dramatic improvements in overall performance as compared to networks that did not incorporate the latent feature and attention framework (e.g., F1 score for presence of heart disease 0.80 vs 0.72). Using cluster assignments and attention based on SHAP values guides deep learning, enhancing latent pattern learning and overall discriminative capability.

## Background

Traditional machine learning and deep learning models have often ignored that not all predictors contribute equally to a specific prediction [1]. However, this perspective is limiting, especially in cases where relationships between predictors and outcomes vary significantly across subgroups. To model these relationships, interaction terms are often relied o n for adapting the behavior of such models, but this too assumes that interactions are considered to have the same effect on every individual. Implicitly, the assumption is that the data are complete, in the sense that all predictors that can affect the model are present in the data, when this is typically not the case for observational data. Thus, a significant challenge in training deep learning models lies in the ability to identify latent structures that underlie the data, as failing to do so limits the model's ability to make accurate predictions. Clustering has been used to group similar observations in an effort to understand these underlying patterns and supervise model training in several clinical and non-clinical domains [2, 3]. These "hybrid" methods aim to leverage the latent structure captured by clustering to alter how the model learns from data, potentially improving predictive accuracy by accounting for subgroup specific behavior. However, neural networks are already capable of learning complex meta features implicitly and thus little can be gained by clustering data in the original feature space. Intrinsically, the reason that features can vary in their predictive importance is due to unobserved features that modify their behavior. To resolve this challenge we propose a method that involves converting the original features into importance scores that reflect their contribution to the outcome of interest and using a suitable clustering technique to gain insights about the underlying subgroups which may have varying amounts of predictive relevance to the outcome of interest. By linking features to outcomes the importance scores do provide information about latent data characteristics Similar approaches though suggested [3], have not been implemented to improve the overall predictive abilities of downstream deep learning models.

SHapley Additive exPlanations (SHAP) [4] is a method in explainable AI that leverages Shapley values from cooperative game theory to fairly allocate contributions of individual features to model predictions. While originally developed for fair allocation in game theory, Shapley values have since been adapted to enhance the interpretability of machine learning and deep learning models by quantifying each feature's contribution to a specific prediction. SHAP values attribute the change in a model's output to individual features, by considering the contribution of each feature across all possible "coalitions" of features; this ensures a fair distribution of feature importance making it a powerful tool for explaining complex models.

In this article, we propose a framework that leverages clustering based on SHAP values and utilizes those cluster identifiers as a feature coupled with an attention mechanism initiated using the SHAP values. This approach makes the resulting clusters independent of the original features themselves and helps to uncover the latent structure. This is because they are disconnected from the feature value itself and instead tell us how much the feature contributes to a model prediction for a specific observation. This implies that similar feature values may contribute differentially to different predictions. This in turn allows us to group observations based on what is predictive for them and not just their original feature values. In fact, this allows us to find latent structures that are related to the outcome by design. Leveraging this information in the overall predictive process can lead to an overall improvement in predictive performance of a deep learning model.

The conceptual advent of neural networks has occurred largely based on attempts to replicate the information processing ability and schema of the human mind. The attention mechanism is one such important concept which tries to simulate how human attention works by assigning varying levels of importance to different features for a given observation based on its outcome [4]. Attention layers initialize weights for each of the features of an observation that are learned during the training process, allowing the model to focus more on the features which have a more important role in a particular outcome. Attention layers have demonstrated improved results in several use cases involving natural language processing [5] and Convolutional Neural Networks [6]. . The representation of each feature is consequently updated based on those feature values whose attention score is the highest. Since SHAP values offer a more refined understanding of feature importance by considering the interaction between features and their individual contributions, using them for initializing the attention weights can potentially provide the model with more accurate starting points for learning. This helps guide the model to start by focusing on features with relatively higher importance, potentially leading to more efficient learning.

In this article, we developed a novel neural network that integrates a SHAP based supervised clustering of data features as well as using these values to initiate an attention mechanism to predict binary outcomes.

## Methods

### Feature-Oriented Representation with Clustering and Explanation (FORCE)

A schematic representation of the FORCE approach is shown in Fig 1.

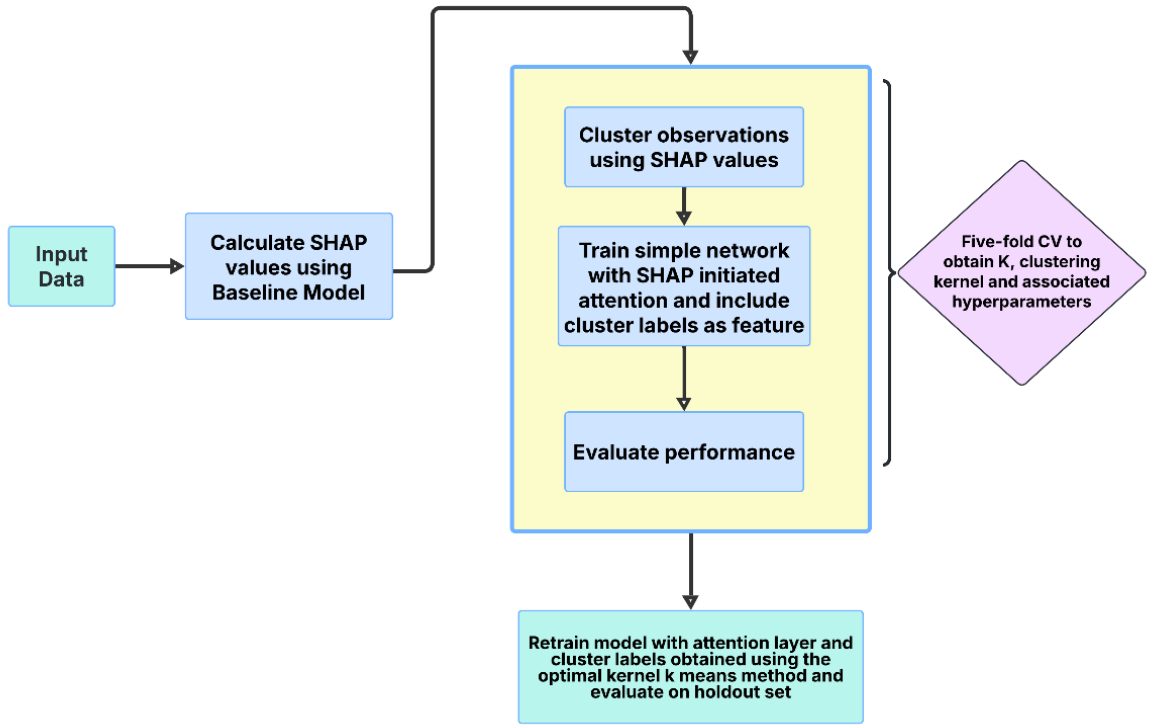

*Figure 1:Schematic of the FORCE (Feature-Oriented Representation with Clustering and Explanation) framework*

FORCE uses three steps:

1. Calculate SHAP values for each observation using a simpler predictive model.
2. Cluster observations using their SHAP values and using the cluster labels as a feature in training the downstream neural network.
3. Use SHAP values to initiate an attention mechanism within a neural network architecture.

### Calculation of SHAP values (step 1)

In this study we utilized a Gradient Boosting Machine (GBM) as a classification model to assess feature importance and baseline predictions. GBM is an ensemble learning technique that builds a strong classifier by sequentially combining multiple weak learners, typically decision trees [7]. In each iteration the model minimizes a differentiable loss function, with the goal of correcting errors made by the prior learners.

We employed the GradientBoostingClassifier in scikit-learn library in python to fit our model on each dataset that underwent transformations as described below in the data description section. The base learners used were decision trees and log loss for binary classification tasks was used.

A random seed was set for reproducibility, and the model's hyperparameters were left to their default settings, including the number of estimators and learning rate. The tree SHAP explainer was used on the feature set and the ensemble model to extract the feature importance values for the model across all the instances.

## Clustering SHAP values (step 2)

The k-means clustering algorithm is an unsupervised learning algorithm that partitions a dataset into a selected number of clusters under some optimization measures such as the sum of squares of Euclidean distances. The assumption behind this method is that the data space can be sub divided into elliptical regions. However, such assumptions may not always hold true, and an alternate idea is to map the data to a new space that satisfies the requirement of the optimization measure. This is achieved by using a kernel function.

The kernel functions allow us to define a transformation to the original data into a new space, allowing us to find clusters that may not be necessarily linearly separable. [8] If we are given a set of samples $x_1,\ldots x_n$ where $x_i \in R^p$ and a mapping function $\phi$ that maps $x_i$ from the input space $R^p$ to the new space $Q$. The kernel function is defined as the dot product in the new space $Q$.

$$H(x_i, x_j) = \phi(x_i) \cdot \phi(x_j)$$

The actual form of $\phi$ is not known and the transformation is defined implicitly. Some common kernel functions follow:

Linear kernel -

$$H(x_i, x_j) = x_i \cdot x_j$$

Polynomial -

$$H(x_i, x_j) = (x_i \cdot x_j + 1)^d$$, where d is the degree of the polynomial space.

Radial -

$$H(x_i, x_j) = \exp(-\gamma \|x_i - x_j\|^2)$$, where $\gamma$ is a tune-able hyperparameter.

The rest of the clustering algorithm is essentially the same as a k-means clustering with appropriate kernels inserted.

The choices of kernel used for this workflow were linear, polynomial, and radial. Cross validation was used to evaluate a range of associate hyperparameters for the polynomial ($d = 2,3$, independent term $= 0,1$) and radial ($\gamma = 0.01, 0.1, 1, 10$) kernels.

## DNN architecture with attention layer initialized by SHAP values (step 3)

SHAP (SHapley Additive exPlanations) values were obtained for all the features using the Gradient Boosting Classifier. These SHAP values were subsequently used downstream to choose the clustering parameters *viz*, kernel type and associated hyperparameters from a five-fold cross validation of the training set. The hyperparameters were selected based on the highest average performance across these metrics from each of the five folds. The SHAP values of the features were used as initiation weights in the attention layer of the neural network. The final clustering hyperparameters for a given dataset was chosen based on the highest values of the average F1-Score.

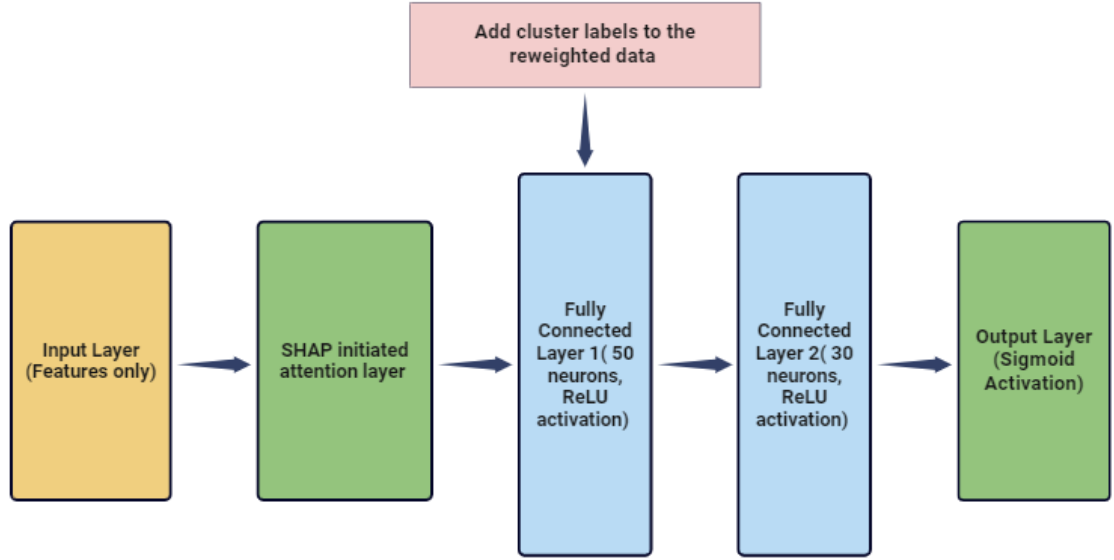

*Figure 2: DNN architecture*

As depicted in Fig. 2, the architecture consists of an attention layer whose weights were initiated using the SHAP values for the observation. The attention layer performed an element-wise multiplication of the sigmoid of weights assigned at an iteration with the feature values. We define the following vectors:

- Feature vector $x = \{x_1, x_2, \ldots, x_n\}$
- Attention weights (initiated by SHAP values) $w = \{w_1, w_2, \ldots, w_n\}$

The attention layer applies a sigmoid function to each weight to normalize it between 0 and 1:

$$\sigma(w_i) = \frac{1}{1 + e^{-w_i}}$$

for each $i = 1, 2, \ldots, n$

Following the sigmoid transformation it performs an element-wise multiplication of the transformed weights and the corresponding feature values, resulting in an attention-weighted feature vector $a$:

$$a_i = \sigma(w_i) x_i$$

for each $i = 1, 2, \ldots, n$

The resulting vector $a = \{a_1, a_2, \ldots, a_n\}$ represents the feature values adjusted by the SHAP initiated attention weights, allowing the model to focus on features with higher weights in each iteration.

The cluster labels were then added to the reweighted features and the concatenated feature set was passed through two fully connected layers having 50 and 30 neurons respectively with ReLU activation. The final fully connected layer feeds into the output layer, which uses a sigmoid activation function for binary classification tasks.

Upon choosing the optimal hyperparameters the model was retrained on the entire training+validation set and evaluated on a holdout test set which was 20% of the entire dataset. The final metrics were compared with a model in which no clustering was performed and no SHAP initiated attention layers were included. The weights were optimized relative to a binary cross-entropy loss function.

The performance of FORCE on each dataset was assessed using accuracy, precision, recall, F1 score, and the area under the ROC curve (AUC) to gain a comprehensive understanding of the method's predictive capabilities. For each of the datasets the metrics were compared to the configuration without including the SHAP cluster labels or the SHAP initiated attention layer. This configuration is referred to as the Simple NN setup from here on.

### Datasets Description

Three datasets were used to train and evaluate the performance of FORCE. These were the (i) Early Stage Diabetes Risk Prediction with 520 observations and 16 features, for which the output variable represents the presence or absence of diabetes [9]; (ii) The Heart disease dataset which has 303 observations and 13 features, the output variable represents the presence or absence of heart disease [10]; and (iii) the credit approval dataset which has 690 observations with 14 features and a status variable indicating whether the observations credit card application was approved or not [11]. All datasets were obtained from the UCI Machine Learning repository.

All datasets underwent minimal preprocessing involving standard scaling of continuous features and one hot encoding of categorical features.

### Results

The performance metrics for each method on all three datasets are shown in Fig. 3. For each of the datasets the model was retrained using the chosen hyperparameters from the five-fold cross validation, incorporating SHAP (SHapley Additive exPlanations) cluster labels into the feature set. An Attention layer for the network was initialized with the SHAP values for the features for each observation. The results for the Simple NN configuration are also presented for comparison. Tables 1-3 present the results of FORCE with two scenarios: (i) randomly initiated attention mechanism and, (ii) using the framework without including the cluster labels as a feature.

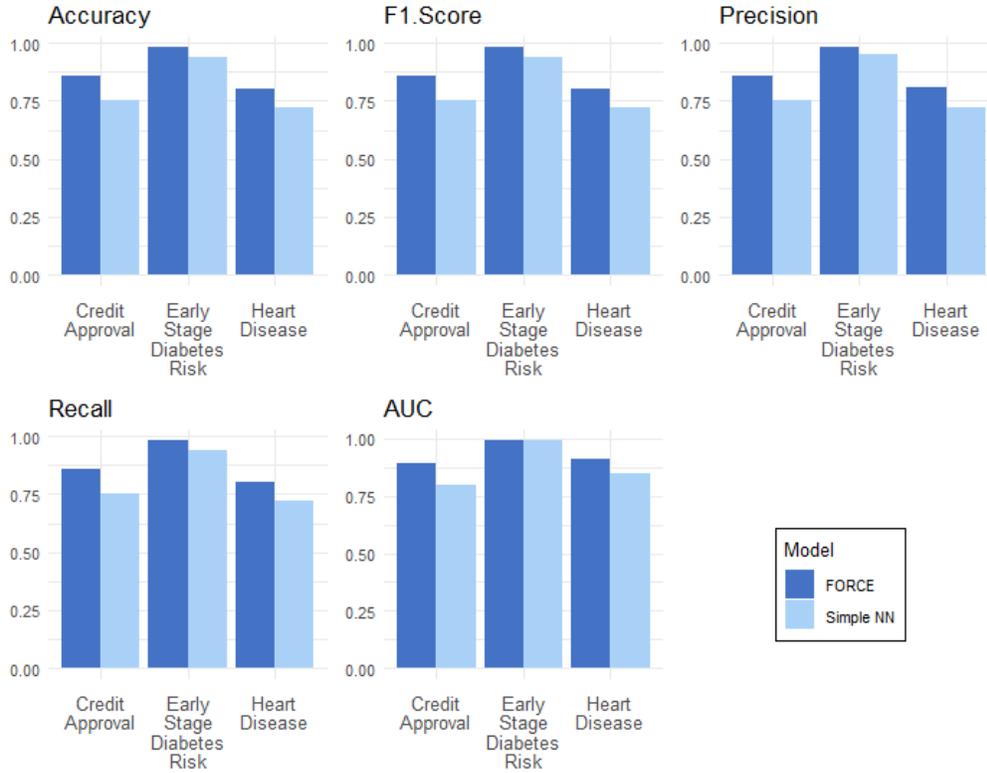

*Figure 3 : Evaluation Metrics for comparing the FORCE to the model without including the cluster labels as a feature and SHAP initiated attention layer*

### Early-Stage Diabetes Risk detection Dataset

The five-fold cross validation for clustering hyperparameters resulted in *k=3* with a third order polynomial kernel and an independent/bias term of 1 for this dataset.

| Metric | FORCE | FORCE with randomly initiated attention | FORCE without cluster labels in feature set |
|---|---|---|---|
| Precision | 0.98 | 0.96 | 0.83 |
| Recall | 0.98 | 0.96 | 0.81 |
| F1-Score | 0.98 | 0.96 | 0.79 |
| Accuracy | 0.98 | 0.96 | 0.81 |
| AUC | 0.99 | 0.99 | 0.92 |

*Table 1: Evaluation metrics of FORCE, FORCE with randomly initiated attention layer, cluster labels as a feature and FORCE with SHAP initiated attention but without including cluster labels into the feature set for the Early Stage Diabetes risk detection data. FORCE outperformed neural networks with randomly initiated attention or without cluster labels on all metrics*

### Heart Disease Dataset

The five-fold cross validation for clustering hyperparameters resulted in k=2 and a linear kernel.

| Metric | FORCE | FORCE with randomly initiated attention | FORCE without cluster labels in feature set |
|---|---|---|---|
| Precision | 0.81 | 0.69 | 0.72 |
| Recall | 0.80 | 0.69 | 0.70 |
| F1-Score | 0.80 | 0.69 | 0.70 |
| Accuracy | 0.80 | 0.69 | 0.70 |
| AUC | 0.91 | 0.80 | 0.74 |

Table 2: Evaluation metrics of FORCE, FORCE with randomly initiated attention layer, cluster labels as a feature and FORCE with SHAP initiated attention but without including cluster labels into the feature set for the Heart disease data. FORCE outperformed neural networks with randomly initiated attention or without cluster labels on all metrics

### Credit Approval Dataset

The five-fold cross validation for clustering hyperparameters resulted in k=5 and a radial basis function kernel with $\gamma = 0.01$ for this dataset.

| Metric | FORCE | FORCE with randomly initiated attention | FORCE without cluster labels in feature set |
|---|---|---|---|
| Precision | 0.86 | 0.79 | 0.83 |
| Recall | 0.86 | 0.79 | 0.82 |
| F1-Score | 0.86 | 0.79 | 0.82 |
| Accuracy | 0.86 | 0.79 | 0.82 |
| AUC | 0.89 | 0.84 | 0.88 |

Table 3: Evaluation metrics of FORCE, FORCE with randomly initiated attention layer, cluster labels as a feature and FORCE with SHAP initiated attention but without including cluster labels into the feature set for the Credit Approval dataset. FORCE outperformed neural networks with randomly initiated attention or without cluster labels on all metrics

### Discussion

In this work, we showed that the Feature-Oriented Representation with Clustering and Explanation (FORCE) method provides a robust and generalizable approach to achieving improved predictive performance across the board. The way FORCE uses SHAP values to guide network architectures in making better predictions stems from an intuitive understanding of SHAP values and their ability to characterize the contribution of features to a particular prediction. The network leverages the inherent properties of SHAP values to characterize the contribution of features to a particular prediction and improve performance. Supervised clustering using SHAP values have been discussed previously as providing clusters that are better separated in lieu of their structured representation of datasets [12]. However, the use of these clustering to capture latent data structure and use it to improve model prediction is an innovative approach. The idea stems from an intuition that if we can allow neural networks to focus not only on supplied features but also on uncovering latent features or characteristics relevant to the outcome, it should enable the network to make more accurate and robust predictions. The use of cluster labels based on clustering SHAP values and initialization of the self-attention mechanism with the observation's SHAP values attains that very objective with improved degree of discrimination as seen in the results.

Although there has been some work in the domain of clustering features and using the attributes of clustering as inputs into various different kinds of network architectures [2, 3], most of the work has focused on using different types of clustering algorithms to cluster the raw values and use them with sophisticated network architectures to gain superior performance. None of these previous works have specifically focused on using latent representations in the data derived from SHAP values which define them based on the importance of features to predictions rather than using the original feature values. The choice of the Gradient Boosting Classifier (GBC) as compared to other models for binary outcomes such as a logistic regression was made because of several advantages that GBC provides over models such as GLMs. These include the ability of GBC in capturing non-linear relationships, automatically accounting for complex feature interactions and providing robust performance even in the presence of multicollinearity. In addition, GBC does not assume linearity or specific distributions, making it better suited to identify underlying patterns in a diverse array of datasets. Thus, SHAP values obtained from GBC give detailed insights into feature importance, which enhances interpretability and helps optimize the prediction process more efficiently.

FORCE uses kernel k -means because it is particularly useful for clustering SHAP values due to its ability to map data into higher dimensional spaces and the flexibility it provides using different kernel functions tailored to the specific characteristics of the data. These properties make kernel k-means especially promising for clustering SHAP values, which inherently represent complex, non-linear relationships between features and model predictions. This ensures that underlying latent structures which provide insights into how features contribute to model behavior become an inherent part of the training process, in turn translating to higher discriminative ability of the model. The presence of tunable parameters in cases of polynomial and radial basis function kernels further allows optimal clustering results for FORCE. These cluster labels, when added as a feature to the prediction job downstream allow the network to learn more about the relative importance of features towards an outcome. In order to understand the contribution of the cluster labels in the prediction we used the same network including the SHAP value-initiated attention layer but without including the cluster labels into the feature set and observed that the presence of cluster labels adds to the predictive performance of the network.

FORCE utilizes an attention layer that is initialized using the observations' SHAP values for the features, which plays a crucial role in focusing the model on the more important features for a given prediction. The attention layer works by providing varying yet randomly assigned weights to the features which are learned as the training progresses, updating weights based on how important a role a feature plays in a particular outcome. Initializing these weights with the observations' SHAP values instead of random initialization reinforces its ability to focus on the important features for a particular prediction, making the process more akin to how human attention works. The SHAP values help in embedding a pre-learned understanding of the feature importance directly into the model architecture. To assess the impact of the SHAP value-initiated attention mechanism, we conducted an experiment where the attention layer was initialized randomly, rather than using SHAP values derived from the observations. This modification resulted in a decline in predictive performance across all three datasets in comparison to FORCE. These results highlight the importance of leveraging features that consistently contribute to outcomes, enhancing the model's ability to generalize effectively to unseen data.

FORCE incorporates SHAP values into the predictive workflow rather than only using them to understand feature contributions after prediction. Our results demonstrate that FORCE provides consistent improvements across all prediction metrics using a simple neural network architecture. The same architecture ends up resulting in worse performance without the integration of the SHAP based implementations. Furthermore, the increase in performance, was consistent across different datasets. The Area Under the curve (AUC) is a threshold independent metric used to assess a models' performance across all possible thresholds We observed that AUC for the FORCE model on the Cleveland Heart

Dataset and the credit card approval dataset increased while it remained the same for the Early Diabetes Detection dataset but essentially it had achieved the nearly perfect performance using the simple network, so no improvement was possible. This indicates that our model is robust across thresholds and has a strong overall predictive power to capture the underlying patterns and relationships in the data that differentiate among the various classes.

While FORCE demonstrates substantial improvement in performance by leveraging SHAP based supervised clustering and self-attention mechanisms, there are potential limitations to be acknowledged within the applicability of the framework. There is a considerable amount of computational complexity involved in calculating SHAP values and performing kernel k-means and this may prove to be a potential limitation in cases with considerably large number of observations. The use of the Gradient Boosting Classifier (GBC) for feature importance analysis was motivated by its ability to capture complex, non-linear relationships and naturally rank features based on their contribution to predictions. However, an integrated approach would allow the importance values to be learned in conjunction with the rest of the model. Future work will explore an integrated model and whether variations in feature importance and interactions influence the overall predictive performance and generalizability of the network. Additionally, although we initialize the attention mechanism with SHAP values, the interpretability of the learned attention weights and their impact on model predictions can often be challenging to understand [13]. The choice of the kernel function varies naturally across datasets based on the hyperparameters. While cross validation is a useful tool in this pursuit it might be of interest to develop specific clustering algorithm for grouping SHAP values with distance functions defined based on their nonlinear and correlated nature. Addressing these limitations in future work will be crucial to enhancing the robustness and applicability of the FORCE framework.

## Conclusion

In this article, we unveil FORCE (Feature-Oriented Representation with Clustering and Explanation) — an innovative framework that combines Shapley Additive Explanations with clustering and attention mechanisms to elevate deep learning model performance. FORCE is designed to uncover hidden patterns and amplify predictive accuracy by intelligently grouping features and focusing attention on what drives predictions for individual observations. Our results demonstrate that this framework consistently improves the overall predictive performance of the model. The improvements are noted for several performance metrics such as accuracy, precision, recall, F1 score and AUC across various datasets, demonstrating the robustness and effectiveness of the framework. The framework not only provides a superior predictive model but also underscores the potential of explainable AI techniques in refining model performance.

## Acknowledgements:

This work was supported by the Kansas Institute of Precision Medicine COBRE P20GM130423 and a CTSA grant from NCATS awarded to the University of Kansas for Frontiers: University of Kansas Clinical and Translational Science Institute (# UL1TR002366). The contents are solely the responsibility of the authors and do not necessarily represent the official views of the NIH or NCATS.

## Data Availability:

All datasets used in the article are publicly available and are obtained from UCI machine learning library.
1. Early Stage Diabetes  Risk prediction dataset :
https://archive.ics.uci.edu/dataset/529/early+stage+diabetes+risk+prediction+dataset
2.  Heart Disease data : https://archive.ics.uci.edu/dataset/45/heart+disease
3. Credit approval data : https://archive.ics.uci.edu/dataset/27/credit+approval

# Supplementary Information

| Metric | FORCE | Simple NN |
|---|---|---|
| Precision | 0.98 | 0.95 |
| Recall | 0.98 | 0.94 |
| F1-Score | 0.98 | 0.94 |
| Accuracy | 0.98 | 0.94 |
| AUC | 0.99 | 0.99 |

Table S1: Evaluation metrics of FORCE and Simple NN for the Early Stage Diabetes Risk data. FORCE outperformed neural networks without incorporating clustering assignments and attention across all metrics.

| Metric | FORCE | Simple NN |
|---|---|---|
| Precision | 0.81 | 0.72 |
| Recall | 0.80 | 0.72 |
| F1-Score | 0.80 | 0.72 |
| Accuracy | 0.80 | 0.72 |
| AUC | 0.91 | 0.85 |

Table S2 : Evaluation metrics of FORCE and Simple NN for the Heart disease dataset. FORCE outperformed neural networks without incorporating clustering assignments and attention across all metrics.

| Metric | FORCE | Simple NN |
|---|---|---|
| Precision | 0.86 | 0.75 |
| Recall | 0.86 | 0.75 |
| F1-Score | 0.86 | 0.75 |
| Accuracy | 0.86 | 0.75 |
| AUC | 0.89 | 0.80 |

Table S3 : Evaluation metrics of FORCE and Simple NN for the Credit approval dataset. FORCE outperformed neural networks without incorporating clustering assignments and attention across all metrics.